\documentclass[sigconf]{aamas} 

\usepackage{balance} 

\setcopyright{none}
\acmConference[OptLearnMAS '23]{Appears in the 14th Workshop on Optimization and Learning in Multiagent Systems (OptLearnMAS 2023)}{May 29, 2023}{London, England}{}
\copyrightyear{2023}
\acmYear{2023}
\acmDOI{}
\acmPrice{}
\acmISBN{}
\acmSubmissionID{???}

\title[Towards Rational Multiagent Systems]{Towards a Unifying Model of Rationality in Multiagent Systems}

\author{Robert Loftin}
\affiliation{
  \institution{Delft University of Technology}
  \city{Delft}
  \country{The Netherlands}}
\email{R.T.Loftin@tudelft.nl}

\author{Mustafa Mert Çelikok}
\affiliation{
  \institution{Aalto University}
  \city{Espoo}
  \country{Finland}}
\email{mustafamert.celikok@aalto.fi}

\author{Frans A. Oliehoek}
\affiliation{
  \institution{Delft University of Technology}
  \city{Delft}
  \country{The Netherlands}}
\email{F.A.Oliehoek@tudelft.nl}

\begin{abstract}
Multiagent systems deployed in the real world need to cooperate with other agents (including humans) nearly as effectively as these agents cooperate with one another.  To design such AI, and provide guarantees of its effectiveness, we need to clearly specify what types of agents our AI must be able to cooperate with.  In this work we propose a generic model of~\textit{socially intelligent agents}, which are individually rational learners that are also able to cooperate with one another (in the sense that their joint behavior is Pareto efficient).  We define rationality in terms of the \textit{regret} incurred by each agent over its lifetime, and show how we can construct socially intelligent agents for different forms of regret.  We then discuss the implications of this model for the development of ``robust'' MAS that can cooperate with a wide variety of socially intelligent agents.
\end{abstract}


\usepackage{tcolorbox}
\usepackage{verbatim}

\DeclareMathOperator*{\argmax}{arg\,max}

\begin{document}


\pagestyle{fancy}
\fancyhead{}


\maketitle 


\section{Introduction}
\label{introduction}

Multiagent systems deployed in the real world (e.g., autonomous vehicle fleets) must be robust to the presence of other intelligent agents that are unlike themselves, such as AI's designed by other companies, or possibly human beings.  Ideally, such systems should not only be robust to other heterogeneous agents, but should be able to reliably cooperate with these agents when it would be mutually beneficial.  The first step in designing such robust, cooperative systems is to identify the types of agents they should be able to cooperate with.  The key challenge is here is that we will often have little or no prior data regarding the behavior of other agents our system might encounter.  The goal of this work is therefore to develop a formal model of \textit{socially intelligent}\footnote{While we borrow the terminology, we cannot claim that our model captures the nuances of the psychological concept of social intelligence~\cite{kihlstrom2000social}.} behavior in multiagent settings, which can serve as the basis for defining robust cooperation, and for creating agents that satisfy this definition.

In developing our social intelligence (SI) model we restrict our attention to the case of two adaptive agents playing a repeated bimatrix game.  Our model imposes two requirements on these agents.  The first is that each be \textit{consistent}, in the sense that they will eventually play a best-response to any fixed partner strategy.  The second is that each agent be \textit{compatible} with some set of agents (potentially just the agent itself), such that when paired with any other member of this set, the joint payoffs will be Pareto efficient (over the set of equilibrium strategies).  Our model is inspired by the \textit{targeted learning} criteria~\cite{powers2004targeted,powers2005finite}, with the key distinction being that we require socially intelligent agents to be consistent against all possible partner strategies.  This implies that socially intelligent agents cannot simply fall back to a non-cooperative, ``safe'' strategy upon encountering another adaptive agent.  In Section~\ref{consistency}, we consider two notions of consistency against non-stationary partners.  Our main theoretical contribution (Section~\ref{socially_intelligent_agents}) is showing how agents satisfying the SI criteria can be constructed for both definitions of consistency.  We conclude with a discussion of the practical utility of this model for the design of multi-agent systems, and pose additional theoretical questions as directions for future work.

\section{Preliminaries}
\label{preliminaries}

We consider the case where interactions between agents take the form of repeated, two-player general-sum matrix games.  We consider classes of games defined by a \textit{type space} $\Theta$, where the current type $\theta \in \Theta$ is known to both agents.  For simplicity, we assume that in all games both players have $N$ pure strategies (henceforth ``actions'') available.  We let $G_1(\theta)$ and $G_2(\theta)$ denote the $N \times N$ payoff matrices for players 1 and 2 in the game defined by type $\theta \in \Theta$.  With a slight abuse of notation, we let $G_i(s_1, s_2; \theta) = s^{\top}_1 G_i(\theta) s_2$ denote the expected payoffs for player $i \in {1, 2}$ under the mixed strategy profile $\langle s_1, s_2 \rangle$.  We assume that agents interact for a fixed number of stages $0 < T < \infty$, and let $a^{1}_t$ and $a^{2}_t$ denote the actions chosen by players 1 and 2 in stage $0 < t \leq T$.  We overload $a^{1}_t$ and $a^{2}_t$ to also denote the mixed strategies that assign all probability mass to actions $a^{1}_t$ and $a^{2}_t$, such that $G_i(a^{1}_t, a^{2}_t; \theta)$ is player $i$'s payoff at stage $t$ given type $\theta$.  We also assume that for all $\theta \in \Theta$ and $a^1 , a^2 \in [N]$, $G_i(a^1, a^2 ; \theta) \in [0, 1]$.  We let $p(s_1, s_2 ; \theta) = \langle G_1(s^1, s^2 ; \theta), G_2(s^1, s^2 ; \theta)\rangle$ denote the players' payoff profile given the joint strategy $\langle s^1, s^2 \rangle$ and type $\theta$.

Let $\mathcal{H}_t = (N \times N)^t$ be the set of histories of play of length $t$ (with $\mathcal{H}_0 = \{ \emptyset \}$), and let $\mathcal{H}_{\leq t} = \bigcup^{t}_{s=0}\mathcal{H}_s$ be the set of all histories of length at most $t$.  An \textit{agent} $\pi$ is pair of mappings $\langle \pi_1, \pi_2 \rangle$ with $\pi_i : \Theta \times \mathcal{H}_{\leq T-1} \mapsto \Delta(N)$, where $\Delta(N)$ is the set of probability distributions over the action set $[N]$.  For each type $\theta \in \Theta$, an agent defines separate \textit{behavioral strategies}~\cite[Chapter~5.2.2]{shoham2008multiagent} for controlling player 1 and 2, which implies that the agent is aware of its own player ID.  Let $h_t \in \mathcal{H}_t$ be the history of play up to stage $t$, and let $s^{i}_t = \pi_i(\theta, h_{t-1})$ be player $i$'s action distribution at $t$, with $a^{i}_t \sim s^{i}_t$.  Each player observes its partner's actions, but not their full action distributions.  We overload $p$ to also denote the empirical average payoff profile for a history $h$, such that
\begin{equation}
    \label{eqn:history_payoff}
    p_i(h; \theta) = \frac{1}{\vert h \vert} \sum^{\vert h \vert}_{t=1} G_i(a^{1}_t(h), a^{2}_t(h) ; \theta),
\end{equation}
where $\vert h \vert$ is the length of the history, and $a^{i}_t(h)$ is player $i$'s action at stage $t$ observed under $h$.  Finally, all probabilities and expectations will be conditional on the current type $\theta$, and the strategies $\pi_1$ and $\pi_2$ selecting actions for players 1 and 2 respectively.  These define a finite probability space over the set of final histories $h_T \in \mathcal{H}_T$.

\section{Consistency and Compatibility}
\label{consistency_and_compatability}

Informally, we say that an agent $\pi$ is \textit{socially intelligent} (SI) if for every type $\theta \in \Theta$ it is both \textit{consistent} with $\theta$ for every possible partner strategy, and \textit{compatible} with itself under $\theta$.  Consistency means that $\pi$ is guaranteed to perform nearly as well as some best-response to its partner's observable behavior, while compatibility means that the joint payoff profile will be nearly as good as that of some Pareto-efficient joint strategy.  In this section, we will consider two formal definitions of consistency, leading to two definitions of social intelligence.

\subsection{Consistent Agents}
\label{consistency}

Our first criteria for social intelligence is that an agent acts as a consistent learner, and attempts to achieve a payoff nearly as large as that of the best response to its partner's strategy.  This is complicated by the fact that the partner's strategy may be non-stationary (particularly if it adapts to the agent as well).  We therefore consider two notions of consistency, \textit{adversarial} and \textit{stochastic}, that account for this non-stationary behavior in different ways.  Our first definition of consistency requires that the agent be robust to such adversarial partners, and relies on the standard notion of \textit{external regret}~\cite{hannan1957consistency}, defined as
\begin{equation}
    \label{eqn:external_regret}
    R^{\text{ext}}_i (h ; \theta) = \!\! \max_{a^{i} \in [N]} \! \sum^{\vert h \vert}_{t=1} \!\! \left\{ G_i(a^{i}, a^{-i}_t (h) ; \theta) - G_i( a^{i}_t (h), a^{-i}_t (h) ; \theta) \right\}
\end{equation}
where $i \in \{1, 2\}$ denotes the player ID of the agent in question, and we use $-i$ to denote the ID of its partner.  Adversarial consistency simply requires that an agent have bounded external regret (i.e., that it be \textit{Hannan consistent}) over $T$ stages.

\begin{definition}[Adversarial Consistency]
\label{def:adversarial_consistency}
For $\delta, \epsilon, T > 0$, an agent $\pi$ is $(\delta, \epsilon, T)$-\textit{adversarially consistent} if, for all types $\theta \in \Theta$, and all partner agents $\tilde{\pi}$, we have that $\frac{1}{T} R^{\text{ext}}_i (h_T ; \theta) \leq \epsilon$ with probability at least $1-\delta$, for either player $i \in \{ 1, 2 \}$.
\end{definition}

While external-regret is theoretically convenient, it is unlikely that most real agents (including humans) would implement a no-regret learning algorithm.  More realistically, we could expect agents to build an empirical model of their partner's behavior, and act optimally with respect to this model.  If the partner could be assumed to choose a fixed strategy $s_{-i}$, then a natural strategy for our agent would be \textit{fictitious play}~\cite{robinson1951fictitious}, under which the agent plays a best-response to its partner's empirical strategy so far.  For history $h$, we define the fictitious play strategy $s^{i}_{\text{fp}}(h; \theta)$ as
\begin{equation}
    \label{eqn:fictitious_play}
    s^{i}_{\text{fp}, t}(h; \theta) \in \argmax_{s \in \Delta(N)} \sum^{t-1}_{r=1} G_i(s, a^{-i}_{r}(h) ; \theta)
\end{equation}
where we choose the $s^{i}_{\text{fp}, t}(h; \theta)$ to be the uniform distribution over all optimal actions given $h_{t-1}$ for player $i$.  We then define the \textit{stochastic regret} $R^{\text{sto}}_i(h ; \theta)$ of player $i$ under history $h \in \mathcal{H}_{\leq T}$ as
\begin{equation}
    \label{eqn:stochastic_regret}
    R^{\text{sto}}_i (h ; \theta) = \sum^{\vert h \vert}_{t=1} \left\{ G_i(s^{i}_{\text{fp}, t}(h; \theta), a^{-i}_t(h) ; \theta) - G_i( a^{i}_t(h), a^{-i}_t(h) ; \theta) \right\}
\end{equation}
This is the difference between the payoff player $i$ would have received had they followed the strategy suggested by fictitious play, rather than $\pi_i$, assuming that player $-i$'s actions would have remained unchanged.  Using this notion of regret, we can now define our second notion of consistency.

\begin{definition}[Stochastic Consistency]
\label{def:stochastic_consistency}
For $\delta, \epsilon, T > 0$, an agent $\pi$ is $(\delta, \epsilon, T)$-\textit{stochastically consistent} if, for all types $\theta \in \Theta$, and all partner agents $\tilde{\pi}$, we have that $\frac{1}{\vert T \vert} R^{\text{sto}}_i (h_T ; \theta) \leq \epsilon$ with probability at least $1-\delta$, for either player $i \in \{ 1, 2 \}$.
\end{definition}

A stochastically consistent agent does not need to be robust to adversarially chosen partner actions.  If changes in the partner's strategy over time would have caused fictitious-play to perform poorly, then a stochastically consistent agent is allowed to perform poorly as well.  The following lemma (proved in Appendix~\ref{apx:regret}) will be useful for our analysis:

\begin{lemma}
\label{lem:regret}
For any history $h \in \mathcal{H}_T$, type $\theta \in \Theta$, and player $i \in \{ 1, 2 \}$, we have that $R^{\text{sto}}_i (h ; \theta) \leq R^{\text{ext}}_i (h ; \theta)$.
\end{lemma}

\noindent We can also define the \textit{expected} adversarial regret as
\begin{equation}
    \label{eqn:expected_external_regret}
    \bar{R}^{\text{ext}}_i (h ; \theta) = \!\! \max_{a^{i} \in [N]} \! \sum^{\vert h \vert}_{t=1} \!\! \left\{ G_i(a^{i}, a^{-i}_t (h) ; \theta) - G_i( s^{i}_t (h), a^{-i}_t (h) ; \theta) \right\}
\end{equation}
and the expected stochastic regret as
\begin{equation}
    \label{eqn:expected_stochastic_regret}
    \bar{R}^{\text{sto}}_i (h_t ; \theta) = \sum^{\vert h \vert}_{t=1} \left\{ G_i(s^{i}_{\text{fp}, t}(h; \theta), a^{-i}_t(h) ; \theta) - G_i( s^{i}_t(h), a^{-i}_t(h) ; \theta) \right\}
\end{equation}
Finally we have
\begin{align}
    \label{eqn:azuma}
    R^{\text{ext}}_i (h_t ; \theta) &\leq \bar{R}^{\text{ext}}_i (h_t ; \theta) + \sqrt{\frac{T}{2} \ln\frac{1}{\delta}}, \\
    R^{\text{sto}}_i (h_t ; \theta) &\leq \bar{R}^{\text{sto}}_i (h_t ; \theta) + \sqrt{\frac{T}{2} \ln\frac{1}{\delta}},
\end{align}
w.p. at least $1-\delta$ for all $t \leq T$ simultaneously (this follows directly from~~\cite[Lemma~4.1]{cesa2006prediction}.  We therefore only need to bound the expected regrets to provide high-probability regret bounds.

\subsection{Compatible Agents}
\label{compatibility}

Consistency captures the idea that an agent is a general-purpose learner.  What makes such a learner socially intelligent, however, is that it is capable of cooperating with other socially intelligent agents.  We therefore need a notion of cooperation that does not preclude consistency.  Let $\mathcal{N}(G) \subseteq \Delta(N) \times \Delta(N)$ be the set of Nash equilibria of the stage game $G$.  For any $s \in \mathcal{N}(G(\theta))$, if both players act according to their respective components of $s$ at each stage, then neither will incur any external (or stochastic) regret in expectation.  For a fully cooperative game $G$ (with $G_1 = G_2$), $\mathcal{N}(G)$ will contain all globally optimal strategy profiles.  It  may, however, also contain strategies that are highly suboptimal, but where neither player can improve the joint payoff by changing their individual strategy, as in the fully cooperative $2 \times 2$ game:
\begin{center}
    \begin{tabular}{c|c|c|}
              & $a^2$ & $b^2$ \\ \hline
        $a^1$ & $2$   & $0$ \\ \hline
        $b^1$ & $0$   & $1$ \\ \hline
    \end{tabular}
\end{center}
As in~\cite{powers2004targeted}, we therefore define compatibility in terms of the \textit{Pareto optimal Nash equilibra} (PONE)~\cite{mas1995microeconomic} of $G$, which we denote as the set $\mathcal{P}(G) \subseteq \mathcal{N}(G)$.  We say that $s \in \mathcal{P}(G)$ if and only if $s \in \mathcal{N}(G)$, and there does not exist $s^{'} \in \mathcal{N}(G)$ such that $G_1(s^{'}) > G_1(s)$ \textit{and} $G_2(s^{'}) > G_2(s)$.  This means that $s$ is a PONE if it is a Nash equilibrium of $G$, and it is not \textit{strongly} Pareto-dominated by any other Nash equilibrium of $G$.  This gives us the following definition of compatibility:

\begin{definition}[Compatibility]
    \label{def:compatibility}
    For $\delta, \epsilon, T > 0$, two agents $\pi$ and $\pi^{'}$ are $(\delta, \epsilon, T)$-\textit{compatible} if, when $\pi_i$ and $\pi^{'}_{-i}$ play together, for any type $\theta \in \Theta$ we have that w.p. at least $1-\delta$, $\exists s \in \mathcal{P}(G(\theta))$ s.t.
    \begin{equation}
         \text{s.t. } p_i(s ; \theta) - p_i(h_T ; \theta) \leq \epsilon,
    \end{equation}
    for either $i = 1$ or $i = 2$.
\end{definition}

A pair of agents is compatible if, when paired together, with high-probability over their path of play $h_T$ there will exist some PONE that does not $\epsilon$-dominate their observed payoff profile $p(h_T ; \theta)$.  Note that this definition of compatibility is very similar to that provided in~\citep{powers2004targeted}, but is now approximate, and defined over a finite time horizon.

\section{Socially Intelligent Agents}
\label{socially_intelligent_agents}

It is natural to model an existing population of agents as a set of compatible, but otherwise heterogeneous agents.  We therefore introduce the more general idea of a socially intelligent \textit{class} of agents that are compatible with any other member of their class:

\begin{definition}
\label{def:social_intelligence}
A set $C$ of agents forms am (adversarially or stochastically) \textit{socially intelligent class} of agents w.r.t. $\Theta$ if, for some $\delta, \epsilon, T > 0$, each agent $\pi \in C$ is (adversarially or stochastically) $(\delta, \epsilon, T)$-consistent for all $\theta \in \Theta$, and any two agents $\pi, \pi' \in C$ are $(\delta, \epsilon, T)$-compatible over $\Theta$.  An individual agent $\pi$ is called \textit{socially intelligent} if it forms a socially intelligent class $\{ \pi \}$ with itself.
\end{definition}

For this notion of social intelligence to be meaningful, it must be possible to construct agents that satisfy the SI criteria.  For both definitions of consistency, we will show that agents using a specific \textit{fallback strategy} satisfy these criteria.  For type space $\Theta$, we first define a function $s(\theta) \in \mathcal{P}(G(\theta))$ that maps from each type $\theta \in \Theta$ to a PONE strategy profile under that type.  We can think of $s(\theta)$ as a ``convention'' the agent or agents have settled upon for the game $G(\theta)$.  Given a consistent agent $\bar{\pi}$, the fallback strategy plays $s_i(\theta)$ at every stage so long as its partner plays the corresponding strategy $s_{-i}(\theta)$.  If its partner eventually fails to play $s_{-i}(\theta)$, the fallback strategy switches to $\bar{\pi}_i$ for all subsequent stages.  

If $s_{-i}(\theta)$ is a mixed strategy, directly testing for deviation from $s_{-i}(\theta)$ is not straightforward.  Instead, the fallback strategy examines the regret the agent has incurred so far, and switches if this exceeds a time-dependent threshold.   As $s(\theta)$ is a Nash equilibrium of $G(\theta)$, we would expect each agent to have small regret when both play according to $s(\theta)$.  Specifically, we have:

\begin{lemma}
    \label{lem:nash}
    For any $\delta, T > 0$, if both players follow strategy $s(\theta)$ at each stage, then then with probability at least $1 - \delta$ we have 
    \begin{equation}
        \bar{R}^{\text{ext}}_i (h_t ; \theta) \leq \sqrt{2T \ln \frac{2}{\delta}}
    \end{equation}
    for all $t \leq T$ and $i \in \{ 1, 2 \}$, and w.p. at lest $1-\delta$ we have
    \begin{equation}
        R^{\text{ext}}_i (h_t ; \theta) \leq 2\sqrt{2T \ln \frac{4}{\delta}}
    \end{equation}
    for all $t \leq T$ $i \in \{ 1, 2 \}$.
\end{lemma}

This follows from an application of the Azuma-Hoeffding inequality (shown in Appendix~\ref{apx:nash}).  Combined with Lemma~\ref{lem:regret} this also provides a bound on the stochastic regret as well.  For both definitions of consistency, we will use Lemma~\ref{lem:nash} to show that the fallback strategy defined by $s(\theta)$ is compatible with itself.

\subsection{Stochastic Social Intelligence}

We first derive a fallback strategy for the case of stochastic regret, which will serve as a template for the adversarial case.  By definition, fictitious play has stochastic regret of zero.  Therefore, the strategy $\pi^{\text{fp}}$ which implements fictitious play (with uniform tie-breaking) for each player is $(\delta, \epsilon, T)$-stochastically consistent for any $\delta, \epsilon, T > 0$.  We define the stochastic fallback strategy $\pi^{T,\epsilon}_i$ for player $i$ as follows:
\begin{enumerate}
    \item While $R^{\text{sto}}_i(h_{t-1} ; \theta) < \epsilon T - 1$, play $s_i(\theta)$.
    \item If $R^{\text{sto}}_i(h_{t-1} ; \theta) \geq \epsilon T - 1$, follow $\pi^{\text{fp}}_i$ for all subsequent stages.
\end{enumerate}

\begin{theorem}
\label{thm:stochastic}
For any $\delta, T > 0$, let $\epsilon_0 \geq 2\sqrt{\frac{2}{T} \ln\frac{4}{\delta}}$, and let $\epsilon = \epsilon_0 + \frac{1}{T}$.  Then the stochastic fallback agent $\pi^{,T,\epsilon}$ is stochastically $(\delta, \epsilon, T)$-socially intelligent.
\end{theorem}

Note that $\pi^{,T,\epsilon}$ will only deviate if $R^{\text{sto}}_i(h_{t-1} ; \theta) \geq \epsilon_0 T$ for some $t \leq T$.  By Lemmas~\ref{lem:regret} and~\ref{lem:nash}, we have that the probability of this happening for either player is at most $\delta$, and so $\pi^{,T,\epsilon}$ is $(\delta, \epsilon_0, T)$-compatible.  As $\pi^{\text{fp}}_i$ will incur no stochastic regret, and since the maximum regret incurred in a single stage is $1$, the maximum possible stochastic regret incurred by $\pi^{T,\epsilon}$ will be $\epsilon T$ surely.  Therefore $\pi^{,T,\epsilon}$ is $(\delta, \epsilon, T)$-stochastically consistent.  Since $\epsilon > \epsilon_0$, $\pi^{,T,\epsilon}$ is also stochastically $(\delta, \epsilon, T)$-SI.

\subsection{Adversarial Social Intelligence}

The case of adversarial regret is somewhat more complex.  Here we base our fallback strategy on the \textit{multiplicative weights}~\cite{freund1999adaptive} update rule, defined as:
\begin{equation}
    s^{i}_{\text{mw},k}(h_t ; \theta) = s^{i}_{\text{mw},k}(h_{t-1} ; \theta) \exp\left(-\eta G_i(k, a^{-i}_{t-1}(h)) \right)
\end{equation}
for $k \in N$, where $s^{i}_{\text{mw}}(h_0 ; \theta)$ is the uniform strategy.  Define $\pi^{\text{mw},T}$ as the agent that plays $s^{i}_{\text{mw}}(h_t ; \theta)$ with learning rate $\eta = \sqrt{8 \ln(N / T)}$.  The expected external regret of $\pi^{\text{mw},T}$ is bounded as
\begin{equation}
    \bar{R}^{\text{ext}}_i (h_T ; \theta) \leq \sqrt{\frac{T}{2}\ln N}
\end{equation}
surely, by~\cite[Theorem 2.2]{cesa2006prediction}.  Similar to the stochastic case, we then define the adversarial fallback strategy $\pi^{T,\epsilon}$ as follows:
\begin{enumerate}
    \item While $\bar{R}^{\text{ext}}_i (h_t ; \theta) \leq \epsilon T - \sqrt{\frac{T}{2}\ln N} - 1$, play $s_i (\theta)$.
    \item Otherwise, switch to $\pi^{\text{mw},T}$ for all subsequent stages.
\end{enumerate}

\begin{theorem}
\label{thm:adversarial}
For any $\delta, T > 0$, let $\epsilon_0 \geq \sqrt{\frac{2}{T} \ln\frac{2}{\delta}}$, and let $\epsilon_1 = \epsilon_0 + \sqrt{\frac{1}{2T} \ln N} + \frac{1}{T}$.  Then for $\epsilon = \epsilon_1 + \sqrt{\frac{T}{2}\ln\frac{1}{\delta}}$, the adversarial fallback agent $\pi^{,T,\epsilon_1}$ is stochastically $(\delta, \epsilon, T)$-socially intelligent.
\end{theorem}

The proof of is similar to that for Theorem~\ref{thm:stochastic}.  By the definition of $\epsilon_1$, $\pi^{,T,\epsilon_1}$ will only deviate when playing with itself if at some point $t \leq T$ one player incurs an expected external regret of at least $\epsilon_0$, and by Lemma~\ref{lem:nash} that will occur with probability at most $\delta$.  Therefore, $\pi^{,T,\epsilon_1}$ is $(\delta, \epsilon_0, T)$-compatible.  We also have that the total expected external regret of the MW agent $\pi^{\text{mw},T}$ is at most $\sqrt{(T / 2) \ln N}$.  This means that if $\pi^{,T,\epsilon_1}$ switches at stage $t$, then the maximum possible expected external regret incurred by $\pi^{,T,\epsilon_1}$ will be less than $\bar{R}^{\text{ext}}_i (h_t ; \theta) + \sqrt{\frac{T}{2}\ln N}$. Since $\pi^{\text{mw},T}$ will always switch just before this point is reached, its total expected regret will be less than $\epsilon_1$ surely, and will be less than $\epsilon$ w.p. $1-\delta$.  As $\epsilon \geq \epsilon_0$, we have that the adversarial fallback strategy $\pi^{,T,\epsilon_1}$ is adversarially $(\delta, \epsilon, T)$-socially intelligent.





\section{Discussion}
\label{discussion}

While we have described a specific approach to designing socially intelligent agents, there are likely many other ways these criteria could be satisfied.  Even restricting ourselves to the fallback strategies considered here, different socially intelligent classes described by different mappings $s(\theta)$ would yield very different behaviors.  A critical theoretical and practical question then is whether we could design a single agent capable of learning to cooperate with any socially intelligent agent.  Under our current definition of social intelligence, this reduces to the problem of learning to cooperate with any consistent agent~\footnote{This is known to be impossible in general (see~\cite{loftin2022impossibility}).}.  To see this, note that for any socially intelligent class $C$, and any arbitrary joint action sequence $\sigma \in \{ N \times N \}^k$, we could construct another class $C'$ that initially play the ``secret code'' sequence $\sigma_i$, and immediately fall back to some arbitrary consistent strategy if the other player fails to do so.  This then raises the related question of how robust a socially intelligent class of agents can be to stochastic or adversarial perturbations of actions taken within an interaction.  It may be possible to establish lower bounds on the probability that cooperation between consistent agents will fail for a given noise distribution.

\section{Related Work}
\label{related_work}

Our model is closely related to the previous targeted learning model~\cite{powers2004targeted,powers2005finite}, which defines similar compatibility and consistency criteria.  The main difference is that targeted learning only requires consistency against a specific target class of partners, which generally would not include the agent itself, or other adaptive agents.  We also require that cooperation and consistent learning occur over a fixed time horizon $T$, rather than asymptotically.  These differences mean that a hypothetical ``universally cooperative'' agent might be able to leverage the consistency of an SI agent to achieve cooperation without a prearranged convention.

This work is partly motivated by the practical challenge of using reinforcement learning to train agents that are able to cooperate with previously unseen partners, a problem sometimes described as \textit{ad hoc teamwork} or \textit{zero-shot coordination}.  A key challenge in using RL for such scenarios is the need to construct populations of training partners (generally trained with RL themselves) that capture the range of cooperative behaviors in the target task~\cite{strouse2021collaborating,cui2021klevel}.  At present, construction of these populations is guided by heuristics that encourage diversity in the strategy space~\cite{lupu2021trajectory,cui2023adversarial,charakorn2023generating}, but do not capture the ability of other agents to adapt to the behavior of others.  By enforcing such a consistency requirement as our SI model does, we would hope to create more realistic training partners for cooperative multiagent RL.

\section{Conclusions}
\label{conclusion}

This work has presented a novel framework for understanding the behavior of rational agents in multiagent scenarios.  We have shown that it is possible to construct classes of consistent learning agents that are also able to reliably cooperate with one another.  Our social intelligence model raises several important theoretical questions that could be explored in future work.  These include determining whether we can design a single agent that can learn to cooperate with any socially intelligent partner, and providing lower bounds on how robust cooperation can be to noisy interactions.  Future work could also consider practical realizations of the social intelligence model for multiagent reinforcement learning, training teams of adaptive agents that (approximately) satisfy the SI criteria.



\begin{acks}
This research was (partially) funded by the Hybrid Intelligence Center, a 10-year programme funded by the Dutch Ministry of Education, Culture and Science through the Netherlands Organisation for Scientific Research, \href{https://hybrid-intelligence-centre.nl}{https://hybrid-intelligence-centre.nl}, grant number 024.004.022.  This work was also supported by the Academy of Finland (Flagship programme: Finnish Center for Artificial Intelligence, FCAI; grants 319264, 313195, 305780, 292334, 328400, 28400) and the Finnish Science Foundation for Technology and Economics (KAUTE).  
\end{acks}



\bibliographystyle{ACM-Reference-Format} 
\bibliography{references}


\section*{Appendices}

\appendix

\section{Proof of Lemma~\ref{lem:regret}}
\label{apx:regret}

We define $P^{a}_i(h; \theta)$ as 
\begin{equation}
    P^{a}_i(h; \theta) = \sum^{\vert h \vert}_{t=1} G_i(a, a^{-i}_t (h) ; \theta),
\end{equation}
and 
\begin{equation}
    P^{\text{ext}}_i(h; \theta) = \max_{a \in N} P^{a}_i(h; \theta).
\end{equation}
We also define $P^{\text{sto}}_i (h ; \theta)$ as
\begin{equation}
    P^{\text{sto}}_i (h ; \theta) = \sum^{\vert h \vert}_{t=1} G_i(s^{i}_{\text{fp}, t}(h; \theta), a^{-i}_t(h) ; \theta)
\end{equation}
We prove by induction on $T$ that $P^{\text{ext}}_i(h_T; \theta) \geq P^{\text{sto}}_i (h ; \theta)$.  This is trivially true for $T = 1$, as $s^{i}_{\text{fp},1}(h_T ; \theta)$ is simply the uniform distribution over all $N$ actions.  Define the regret of fictitious play w.r.t. action $a \in N$ as
\begin{equation}
    R^{a}_i(h; \theta) = P^{a}_i(h; \theta) - P^{\text{sto}}_i (h ; \theta)
\end{equation}
and the external regret of fictitious play $R_i(h; \theta)$ as
\begin{equation}
    R_i(h; \theta) = P^{\text{ext}}_i(h; \theta) - P^{\text{sto}}_i (h ; \theta).
\end{equation}
We also define the set $A_i (h_ ; \theta)$ as
\begin{equation}
    A_i (h ; \theta) = \argmax_{a \in N} R^{a}_i(h; \theta).
\end{equation}
We observe that
\begin{align}
    R_i(h_{T + 1}; \theta) &= \max_{a \in N} \left\{ G_{i}(a, a^{-i}_{T+1} (h_{T+1}); \theta) \right. \\ 
    &\left. - G_{i}(s^{i}_{\text{fp},T+1}(h_{T+1} ; \theta), a^{-i}_T (h); \theta) +  R^{a}_i(h_T; \theta) \right\}
\end{align}
By the definition of fictitious play, we have that for any action $a \in A_i (h_{T} ; \theta)$, the probability of $a$ under $(s^{i}_{\text{fp},T+1}(h_{T+1} ; \theta)$ is $> 0$.  Assuming that $R_i(h_T; \theta) \geq 0$, this means that for all $a \in A_i (h ; \theta)$, $R^{a}_i (h_T; \theta) \geq 0$, and there exists $a' A_i (h_{T} ; \theta)$ such that
\begin{equation}
    G_{i}(a', a^{-i}_{T+1} (h_{T+1}); \theta) - G_{i}(s^{i}_{\text{fp},T+1}(h_{T+1} ; \theta), a^{-i}_T (h); \theta) \geq 0.
\end{equation}
This in implies that $R^{a'}_{i}(h_{T + 1}; \theta) \geq 0$, which in turn implies that $R_{i}(h_{T + 1}; \theta) \geq 0$.  This means that the payoff of the best action in hindsight is always at least as large as the accumulated payoff of fictitious play, and proves Lemma~\ref{lem:regret}. $\square$

\section{Proof of Lemma~\ref{lem:nash}}
\label{apx:nash}

Here the type $\theta$ will be implicit.  For $i \in \{ 1, 2\}$, we define $V^{i}_t$ as
\begin{equation}
    V^{i}_t = G_i (s^{i}_t , s^{-i}_t) - G_i (s^{i}_t, a^{-i}_t)
\end{equation}
We can see that $\text{E}[V^{i}_t \vert h_{t-1}] = 0$.  We can then have that
\begin{align}
    \bar{R}^{\text{ext}}_t &= \max_{a \in N} \sum^{t}_{r=1} \left\{ G_i(a, s^{-i}_r) - G_i (s^{i}_r, a^{-i}_r) \right\} \\
    &= \max_{a \in N} \sum^{t}_{r=1} \left\{ G_i(a, s^{-i}_r) - G_i (s^{i}_r , s^{-i}_r) + G_i (s^{i}_r , s^{-i}_r) - G_i (s^{i}_r, a^{-i}_r) \right\} \\
    &= \sum^{t}_{r=1} \left\{ G_i (s^{i}_r , s^{-i}_r) - G_i (s^{i}_r, a^{-i}_r) \right\} = \sum^{t}_{r=1} V^{i}_r \\
    &\leq \sqrt{\frac{2}{T}\ln\frac{1}{\delta}}
\end{align}
with probability $1-\delta$ for all $t \leq T$ simultaneously.

This follows from the fact that $\vert V^{i}_t \vert \in [0,1]$ and the ``maximal'' Azuma-Hoeffding inequality~\cite{hoeffding1963probability}.  The second equality follows from the fact that $\langle s^{i}_t , s^{-i}_t \rangle = s(\theta)$ is a Nash equilibrium.  The first bound of Lemma~\ref{lem:nash} follows from a union bound over the probability for both players, while the second bound combines this with Equation~\ref{eqn:azuma}. $\square$


\end{document}